\documentclass[conference]{IEEEtran}  

\IEEEoverridecommandlockouts                              

\usepackage{graphicx} 
\usepackage{caption}
\usepackage[caption=false,font=footnotesize]{subfig}
\usepackage{amsmath} 
\usepackage{amssymb}  
\usepackage{tikz}
\usepackage{pgfplots}
\pgfplotsset{compat=1.18}
\usepackage{siunitx}
\usepackage{booktabs}
\usepackage{bm}
\usepackage{hyperref}
\usepackage{rotating}
\usepackage{cite}
\usepgfplotslibrary{groupplots}
\usepackage[nolist]{acronym}
\usetikzlibrary{arrows.meta, positioning, shapes.geometric, shapes.symbols,
                 calc, fit, backgrounds, decorations.markings, shadows.blur}
\usepackage{xcolor}

\definecolor{eventblue}{RGB}{41,98,163}
\definecolor{procgreen}{RGB}{39,139,89}
\definecolor{measorange}{RGB}{217,119,39}
\definecolor{ekfpurple}{RGB}{118,55,154}
\definecolor{outputred}{RGB}{186,50,50}
\definecolor{lightbg}{RGB}{245,247,250}
\definecolor{stagegray}{RGB}{230,233,238}

\definecolor{caruco}{HTML}{009E73}
\definecolor{castratag}{HTML}{0072B2}
\definecolor{capriltag}{HTML}{D55E00}
\definecolor{ctauA}{HTML}{0072B2}
\definecolor{ctauB}{HTML}{E69F00}
\definecolor{ctauC}{HTML}{CC79A7}

\pgfplotsset{
    draxis/.style={
        ymin=0, ymax=105, ytick={0,20,40,60,80,100},
        grid=both, grid style={gray!18, line width=0.3pt},
        tick align=outside, tick pos=left,
        label style={font=\small}, tick label style={font=\footnotesize},
        title style={font=\small, yshift=-1ex}, legend cell align=left,
    },
    aruco/.style   ={caruco,    mark=*,         line width=1pt, mark size=2pt,  mark options={fill=white, draw=caruco,    line width=0.8pt}},
    astratag/.style={castratag, mark=square*,   line width=1pt, mark size=2pt,  mark options={fill=white, draw=castratag, line width=0.8pt}},
    apriltag/.style={capriltag, mark=triangle*, line width=1pt, mark size=2.4pt,mark options={fill=white, draw=capriltag, line width=0.8pt}},
    tauA/.style={ctauA, mark=square*,   line width=1pt, mark size=2pt,  mark options={fill=white, draw=ctauA, line width=0.8pt}},
    tauB/.style={ctauB, mark=*,         line width=1pt, mark size=2pt,  mark options={fill=white, draw=ctauB, line width=0.8pt}, densely dashed},
    tauC/.style={ctauC, mark=triangle*, line width=1pt, mark size=2.4pt,mark options={fill=white, draw=ctauC, line width=0.8pt}, densely dotted},
}

\acrodef{RPOD}[RPOD]{Rendezvous, Proximity Operations, and Docking}
\acrodef{ICP}[ICP]{Iterative Closest Point}
\acrodef{TPS}[TPS]{Thin Plate Splines}
\acrodef{ISAM}[ISAM]{In-Space Assembly and Manufacturing}
\acrodef{OOS}[OOS]{On-Orbit Service}
\acrodef{ADR}[ADR]{Active Debris Removal}
\acrodef{RSO}[RSO]{Resident Space Object}
\acrodef{GRS}[GRS]{Generalised Reed–Solomon}
\acrodef{AIAA}[AIAA]{American Institute of Aeronautics and Astronautics}
\acrodef{ESA}[ESA]{European Space Agency}
\acrodef{ISRO}[ISRO]{Indian Space Research Organization}

\acrodef{CONFERS}[CONFERS]{The Consortium for Execution of Rendezvous and Servicing Operations}
\acrodef{CLAHE}[CLAHE]{Contrast-Limited Adaptive Histogram Equalization}

\title{\LARGE \bf
Spacecraft Fiducial Marker for Autonomous Rendezvous, Proximity Operations, and Docking
 }

\author{
\IEEEauthorblockN{
Ravi Kumar Thakur*, Matouš Vrba, Martin Saska}

\thanks{Authors are with the Department of Cybernetics,  Faculty of Electrical Engineering, Czech Technical University in Prague, Karlovo Namesti 13, 121 35 Prague 2, Czechia. *Corresponding author: {\tt\small ravi.thakur@cvut.cz}}%

}

\begin{document}

\maketitle
\thispagestyle{empty}
\pagestyle{empty}

\begin{abstract}

Robotic operations in space are challenging due to the harsh environment and the high cost of failure. Fiducial markers provide visual references that aid autonomous rendezvous, proximity operations, and docking for space robots. However, existing fiducial markers are mostly single scale , largely designed for terrestrial robotics. Such markers leave the camera's field of view at close range, precisely during the proximity and docking phases where reliable tracking is most critical. This paper presents AstraTag, a fiducial marker designed for autonomous on-orbit robotic operations. The marker template is based on a square Spidron pattern whose recursive, self-similar structure enables detection across multiple spatial scales. Marker identification uses a 48-bit signature derived from triangular sub-regions of the template and encoded with a Generalised Reed–Solomon (GRS) code. The detection pipeline performs contour-based quadrilateral localisation, perspective normalisation, and signature matching against a pre-computed dictionary. To handle markers affixed to curved spacecraft surfaces, it incorporates a Thin-Plate Spline (TPS) re-warp fallback that exploits the marker's internal rectangular borders as additional geometric correspondences. We benchmark AstraTag against three-layer Fractal ArUco and AprilTag on spacecraft mockups with flat and curved surfaces. On curved surfaces, AstraTag achieves a higher detection rate than both baselines, offering a robust recursive-marker option for space robotics. 

\end{abstract}


\section{INTRODUCTION}

\begin{figure}[t]
    \centering
    \includegraphics[width=\columnwidth]{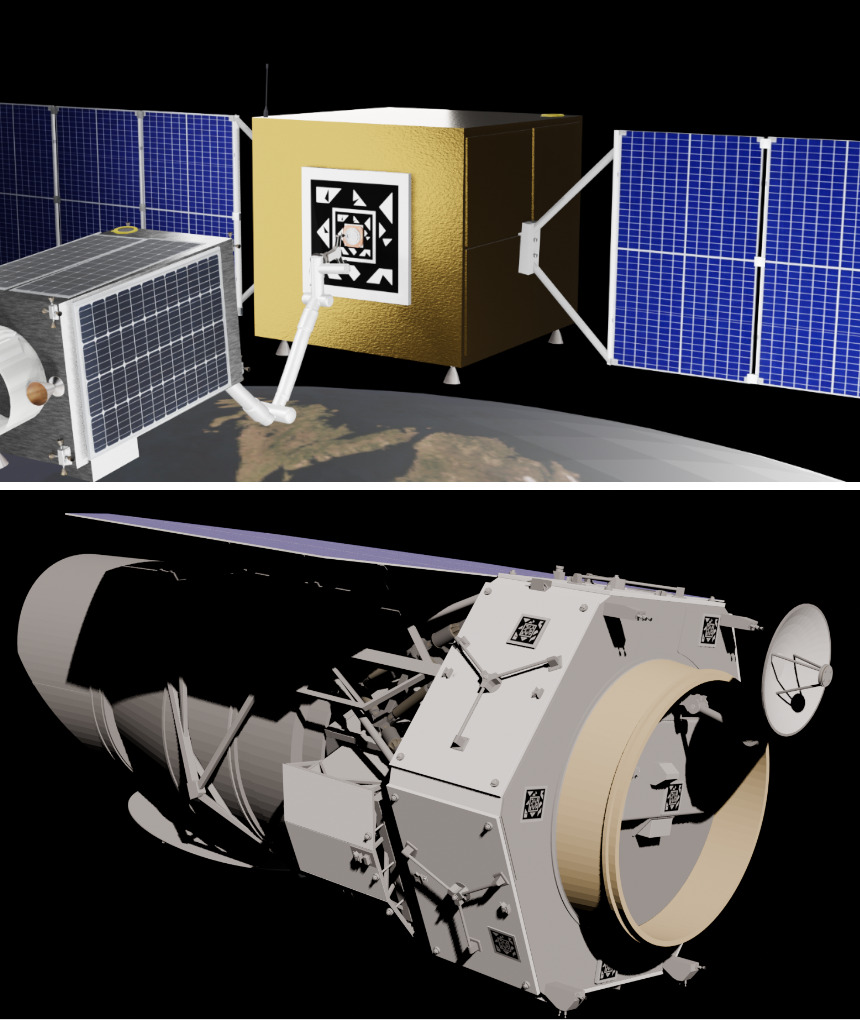}
    \caption{The AstraTag marker in an illustration of on-orbit robotic operations. The top image shows a robotic servicer spacecraft performing proximity around target docking port for refueling. The bottom image~(\textit{Credits: NASA}) shows the model of Nancy Grace Roman Telescope with AstraTag as potential servicing fiducials on its spacecraft bus.}
    \label{fig:intro}
\end{figure}

Autonomous robotic operations are crucial for future space missions. With continuously increasing number of \acp{RSO}, the scope of space activities has increased with the requirement of sustainable space missions. It has become necessary to work on and around such \ac{RSO} irrespective of whether they are cooperative or non-cooperative. The objective may include capability for \ac{ISAM} for the construction of a larger space structure, \ac{OOS} for spacecraft maintenance and refueling, and \ac{ADR} missions to de-orbit the spacecraft post-mission life-cycle. If unsuccessful, each of these occurrences can result in collisions that result in space debris. A common denominator for all these missions is \ac{RPOD}. Conducting RPOD autonomously requires an accurate relative state estimate of the Chaser and the Target spacecraft as shown in Figure~\ref{fig:intro}.

The current approaches for \ac{RPOD} for critical missions involve semi-automated procedures with human-in-the-loop teleoperation. To enable autonomous \ac{RPOD}, spacecrafts need to have an accurate estimate of relative states to aid their guidance, navigation, and control system. However, spacecrafts operate in the harshest environment under the most conservative size, weight, and power constraints. The onboard autonomy and intelligence is limited both by the software and the hardware. Thus, more advanced markerless vision-based techniques for vision based relative state estimation can add to the complexity of already challenging \ac{RPOD} operations, since estimating the six degree-of-freedom pose estimate from visual features alone requires information-rich images. Space environments are, in general, perceptually challenging due to the extreme contrast between shadows and illuminated surfaces, strong reflections, and bright light sources combined with a dark background.





\begin{figure*}[t]
    \centering

        \includegraphics[width=\textwidth]{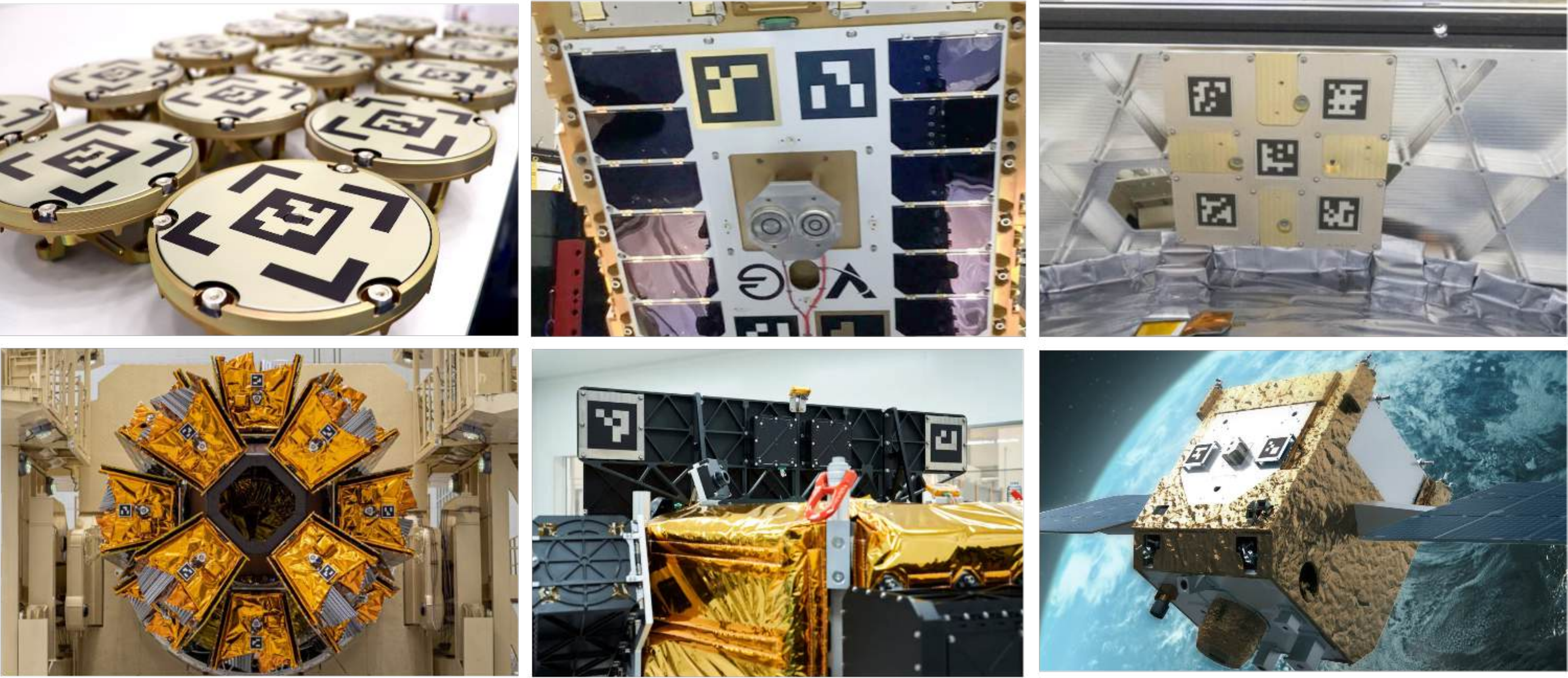}
    \caption{Examples of fiducial-based systems for in-space proximity operations (clockwise from top-left): Docking Plate~(\textit{Astroscale}) ,  RAFTI refueller~(\textit{Orbit Fab}), Servicing Fiducials~(\textit{GSFC, NASA}), Provisioner refueller ~(\textit{Astroscale}), Servicing Fiducial~(\textit{Space Machines Company}), DogTag installed on OneWeb satellite~(\textit{Voyager and Altius Space Machines}).}
    \label{fig:space_applications}
\end{figure*}

In such scenarios, fiducial markers provide robust artificial landmarks for detection and pose computation. These markers are mature and can complement markerless relative state estimation algorithms onboard the spacecraft. A fiducial marker is an object with known geometry and texture placed in the environment as a visual reference. Its design is driven by application requirements: some require robust detection under degraded imaging conditions, while others prioritize error-free identification or accurate pose estimation. Fiducial markers have been employed in several space applications, including the Space Vision Marker System (SVMS) used during Shuttle--ISS rendezvous~\cite{bondy2007space}, NASA's Astrobee free-flying robot on the International Space Station~\cite{smith2016astrobee}, or the rendezvous fiducial implementations for servicing missions~\cite{skelton2024rendezvous}. Altius Space Machines developed DogTag as grappling fixture for small satellite systems for active debris removal missions~\cite{maclay2020development}. Figure~\ref{fig:space_applications} shows the existing use cases of the fiducial markers in space missions. 

The growing importance of standardised fiducial markers for space applications is reflected in the draft developed by the \ac{AIAA} On-Orbit Servicing and Assembly Committee in collaboration with \ac{CONFERS}~\cite{aiaa_s155}. The standard defines baseline functional, physical, and operational requirements for high-contrast spacecraft fiducial markers, such as, unique identification, crisp edges, and six degrees-of-freedom pose estimation, establishing a minimum capability floor for the industry. The standard takes into account current practices and anticipates several challenges that the next generation of on-orbit servicing will demand. Our proposed fiducial marker, AstraTag, is a contribution in that direction. It is designed for close range detection on curved and non-planar surfaces, resilience across wide distance ranges from a single marker, graceful degradation under partial occlusion or illumination damage, and dictionary sizes that scale to large constellations.

The proposed marker AstraTag is based on a square Spidron~\cite{erdely2008some} pattern, a self-similar geometric structure composed of alternating triangles. The Spidron template enables multi-range detection: the same encoded pattern is replicated at the center of the marker at a reduced scale, so that when the outer layer becomes too small to resolve at long range, the inner layer may still be detectable at close range. Each triangular region of the Spidron is subdivided into six sub-triangles by connecting the three vertices, three edge midpoints, and an approximate centroid, yielding forty eight encoding cells across all eight regions. The resulting binary signature is protected by a \ac{GRS}~\cite{shokrollahi2013class} that provides error correction against local image degradation (see Figure~\ref{fig:marker_template}). AstraTag's detection pipeline incorporates a \ac{TPS} re-warp fallback using \ac{TPS} implementation that uses the marker's internal white rectangular borders as additional correspondences to correct the non-linear distortion\cite{bookstein1989principal}. AstraTag is compared with existing recursive fiducial markers in real-world experiments on both the flat and curved surfaces of mockup spacecraft in environments mimicking the lighting conditions of space.

The contributions of this paper are as follows.
\begin{itemize}
    \item Design of a recursive fiducial marker based on the Spidron fractal pattern with \ac{GRS}-based encoding and area-based signature sampling.
    \item A detection pipeline that operates across varying distances via the recursive multi-scale structure.
    \item Integration of \ac{TPS}-based unwarping using the marker's internal rectangular borders as geometric correspondences, extending reliable detection to 70\textdegree{} out-of-plane rotation on curved surfaces.
    \item A comparative benchmark against Fractal ArUco and AprilTag on two datasets, demonstrating AstraTag's advantage on curved spacecraft surfaces.
\end{itemize}

The source code is available at \url{https://github.com/astradyn/astratag}.

\section{RELATED WORK}

Most fiducial markers in robotic applications use bi-tonal black-and-white patterns for encoding. AprilTag\cite{krogius2019flexible}, ARTag\cite{fiala2005artag} and ArUco\cite{munoz2020ucoslam} are some of the widely used ones in this category. AprilTag was designed for use with an autonomous robotic systems. Although ArUco and ARTag are similar in design, they provide options for different signature resolutions. These markers are detected through contour approximation and the checkerboard pattern inside is used to encode the identification. Consequently, these markers need to be printed in larger sizes for detection and identification and use the outer contour for pose estimation. The PiTag marker utilizing a pattern of dots positioned in a square shape was proposed to address occlusion by introducing computation in the image space by\cite{bergamasco2013pi}. A derivative of Pi-Tag with a circular shape was proposed as Rune Tag\cite{bergamasco2011rune}.  This marker has dots placed along the invisible circumference. Although the dot patterns make these markers robust to occlusions, they are computationally expensive to detect.

There are a few markers with varying contrast instead of bi-tonal patterns. Fourier Tag\cite{xu2011fourier} was designed for underwater detection and relies on a grayscale texture inside a circular disk. The identity in this marker is encoded using the Fourier transform. Chroma Tag was proposed as an alternative to binary and grayscale markers by making colorful April Tags\cite{degol2017chromatag}. The use of feature descriptors like SIFT and SURF can be used for maximum detection in a blob-like marker\cite{schweiger2009maximum}. However, the size of such marker dictionary is limited. In \cite{zhang2022deeptag}, the Deep Tag was proposed as a deep learning-based method for marker detection and identification. The model is trained on all possible fiducial markers available and performs detection and matching using this trained model.  


Robotic operations in space can face challenging perceptual conditions due to solar illumination, shadow, and reflection from spacecraft surfaces~\cite{amaya2024visual}. For such scenarios, the fiducial marker must offer detection at various ranges and be robust against degradation, to allow \ac{RPOD} operations within a wider approach corridor. To allow detection on multiple scales, Fractal Markers~\cite{romero2019fractal} introduced a recursive design by nesting ArUco markers at multiple concentric levels. Each level carries a different ArUco code, enabling trivial layer identification and graceful degradation under occlusion. However, the use of different codes per layer requires larger dictionaries, and the square grid encoding limits the information density within each layer. AprilTag~\cite{krogius2019flexible} introduced flexible layouts that allow multiple markers to be tiled. In both cases, the outer quadrilateral contour provides the four correspondences used for pose estimation. Similar approaches to create multi-layer markers were also proposed in \cite{herout2012fractal}, \cite{tybusch2017color}, \cite{wang2018hierarchical}. A circular-shaped binary marker named CCTag was proposed in \cite{gatrell1992robust}, which consists of concentric contrasting disks with sharp intensity transitions at the boundaries. The marker was designed for robust tracking for space applications. However, the pose estimation required multiple sets of this marker optimally placed on large surface area of the spacecraft.

\section{SPACECRAFT FIDUCIAL MARKER}

\begin{figure}[t]
\includegraphics[width=\columnwidth]{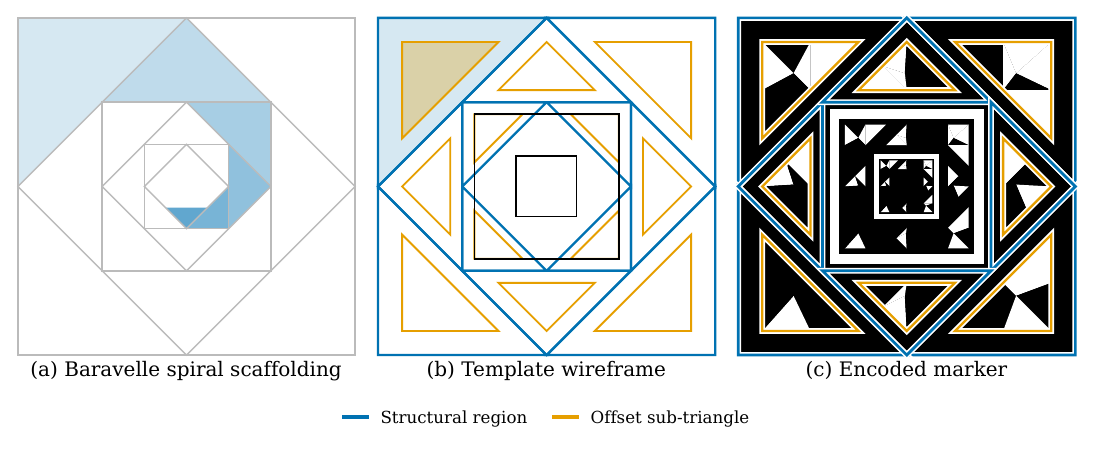}
\caption{(a)~The square Spidron pattern formed by recursively nesting rotated squares, with one triangular region highlighted in each layer. (b)~The AstraTag marker template derived from the square Spidron: the wireframe showing the twelve structural regions and offset boundaries (left), and a marker after triangular subdivision and binary encoding (right).}
\label{fig:marker_template}
\end{figure}

\subsection{AstraTag}
The AstraTag marker template is based on a square Spidron, which is a group of self-similar fractal geometric patterns made of alternating sequences of triangles\cite{erdely2008some}. The AstraTag template adapts the Spidron concept to a square perimeter, inspired by square Spidron diagrams from the Irish Mathematical Trust\cite{irishmathstrust} (see Figure~\ref{fig:marker_template}a). The square boundary was chosen for the utility of the outer quadrilateral contour in perspective-$n$-point pose estimation~\cite{haralick1994review}. The square Spidron template is similar to a Baravelle spiral generated by a square to create geometric visualization of infinite series\cite{harper2000exploring}. The scaffolding of the AstraTag template as shown in Figure~\ref{fig:marker_template}) is built by recursively inscribing a polygon in the side-midpoints of the previous
one. For a regular $n$-gon, this midpoint step produces a similar $n$-gon scaled by the factor
\begin{equation}
    \sigma_n = \cos\!\left(\frac{\pi}{n}\right)
    \label{eq:scale}
\end{equation}
and rotated by the angle $\varphi_n = \pi/n$, where $n=4$ in the case of AstraTag. This gives $\sigma_4 = \cos(\pi/4) = 1/\sqrt{2}$ and $\varphi_4 = \pi/4$. Thus, each nested square has half the area of its parent and is turned by $\pi/4$.
Indexing the nesting level by $k=0,1,2,\dots$, the side length and area of the
$k$-th square therefore obey
\begin{equation}
    s_k = s_0\,2^{-k/2},\qquad A_k = s_k^{2} = A_0\,2^{-k},
    \label{eq:nesting_law}
\end{equation}
where $s_0$ and $A_0$ are the side length and area of the outer square ($k=0$),
and $s_k$, $A_k$ are those of the $k$-th square.





The recursive property of the Spidron structure is exploited by replicating the outer encoding pattern at the centre of the marker at a reduced scale. The four central regions provide structural symmetry but do not carry encoded information. To prevent encoded cells from merging with the marker boundary or neighboring regions during detection, each structural triangle is contracted toward its centroid by a fixed factor. The marker comprises three concentric layers. The white rectangular borders separate adjacent layers and serve as structural features for both layer delineation and curved surface correction. This separation improves quadrilateral detection to locate the marker. The marker design can be parameterized using the offset and border thickness for a desired detection range.

\subsection{Encoding}
\begin{figure}[t]
\centerline{\includegraphics[width=\columnwidth]{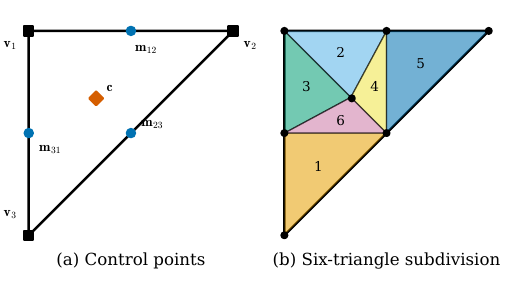}}
\caption{Triangular subdivision of one Spidron encoding region. (a)~The seven control points: three vertices (black squares), three edge midpoints (blue circles), and the approximate centroid (red diamond). (b)~The resulting six-triangle subdivision used for encoding.}
 \label{fig:trisampling}
\end{figure}

The triangular subdivision uses seven control points derived from the offset sub-triangle of each region, the three vertices, three edge midpoints, and the centroid. These seven points are connected into six non-overlapping triangles that tile the region. This results in total 48 encoding cells. The subdivision is computed once and stored as a keypoint file shared by both the marker generator and the detector. Figure~\ref{fig:trisampling} illustrates the seven control points and the resulting six-triangle subdivision for one encoding region.

To protect against bit errors caused by image degradation, partial occlusion, or imperfect binarisation, the marker signature is structured as a \ac{GRS} codeword. The values from this codeword are used to selectively solid-fill the 48 triangular cells. Each triangle is assigned either a white or black color. The signature is extracted for all the four orientations of the marker under cyclic rotation. For each marker identity, the dictionary stores all four oriented signatures together with the corresponding world corner coordinates. Storing all orientations explicitly trades dictionary size and safety of docking procedure for detection speed. This enables co-operative docking only when the servicing spacecraft has access to the dictionary corresponding to the markers in target spacecraft.

\subsection{Detection}
The AstraTag detection pipeline operates in two stages, quadrilateral localisation and signature-based identification. The raw grayscale image is processed with \ac{CLAHE}\cite{zuiderveld1994contrast} to normalise local contrast and improve edge visibility under uneven illumination. The resulting image is converted to a binary image using a gaussian adaptive threshold which computes a local threshold for each pixel neighborhood, making binarisation robust to illumination gradients and
partial shadows.

\subsubsection{Quadrilateral Detection}
The quadrilateral boundary is detected by extracting contours from the binary image using hierarchical retrieval, which separates outer boundaries from inner holes and allows the nested structure of the recursive marker to be represented in the contour tree. Contours with an area below a threshold are discarded. The remaining contours are binarized and analyzed using the Probabilistic Hough Transform with parameters to detect the constituent line segments. The adjacent segments are grouped into parallel pairs. A valid candidate must contain at least one pair of mutually parallel sides. This line-based validation recovers plausible quadrilaterals even when contour closure is incomplete due to partial occlusion or local degradation. The detected candidates are refined using polygon approximation~\cite{douglas1973algorithms} to obtain exactly four corner vertices. The candidate quadrilaterals are further filtered based on aspect ratio. The corner coordinates are then refined to sub-pixel accuracy using gradient-based fitting for accurate pose estimation.

\subsubsection{Signature Extraction}

The identity of the marker is computed after the quadrilateral is confirmed.  A perspective homography maps the four detected corners to a view of fixed pixel dimensions. The warped patch is further binarised to account for residual illumination variation. Each signature bit is then determined by area-based sampling over the corresponding triangular cell. This area-based approach integrates hundreds to thousands of pixels per cell, making the signature robust to localized noise and illumination gradients. The observed signature is compared with all dictionary entries using the Hamming distance. A detection is accepted if the distance computed is below the maximum allowed Hamming distance. A \ac{TPS}-based curved surface fallback is applied when the standard pipeline fails on non-planar surfaces. The pipeline is designed to be robust against varying illumination, out-of-plane rotation, and scale changes. 





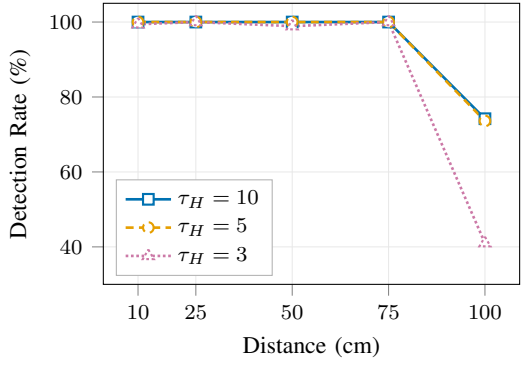
\begin{figure}[t]
    \centering
    \begin{tikzpicture}
    \begin{axis}[draxis, width=0.8\linewidth, height=5.3cm,
        xlabel={Distance (cm)}, ylabel={Detection Rate (\%)},
        xtick={10,25,50,75,100}, ymin=30, ymax=105,
        legend style={at={(0.03,0.03)}, anchor=south west, font=\footnotesize,
            fill=white, draw=gray!60}]
        \addplot[tauA] coordinates {(10,100)(25,100)(50,100)(75,100)(100,74.2)};
        \addplot[tauB] coordinates {(10,100)(25,100)(50,100)(75,100)(100,73.6)};
        \addplot[tauC] coordinates {(10,99.4)(25,100)(50,98.9)(75,100)(100,41.0)};
        \legend{$\tau_H=10$, $\tau_H=5$, $\tau_H=3$}
    \end{axis}
    \end{tikzpicture}
    \caption{Real-world evaluation of AstraTag detection recall over distance of the target from a $320\times240$ resolution camera for different thresholds of the Hamming error. }
    \label{fig:multi_range}
\end{figure}

The marker encodes information at multiple spatial scales through its recursive structure. At a larger range, the outer layer occupies a large portion of the image and is detected first. At closer distances, the outer layer may become too small to resolve.  In such cases, the inner sub-marker centered within the marker serves as the detection target. The pipeline applies the same quadrilateral detection and signature extraction independently to both both  the outer and the inner regions, enabling detection across a wider approach corridor than a single-scale marker would allow. 



\subsection{Pose Estimation}
The pose computation in AstraTag uses Infinitesimal Plane-based Pose Estimation (IPPE)~\cite{collins2014infinitesimal}, an analytical method for planar targets. The detected corners of marker boundary $\mathbf{u}_i$ are matched to their known three-dimensional world points stored in the dictionary. The pose satisfies the perspective projection,

\begin{equation}
    \lambda_i\,{\mathbf{u}}_i
      = \mathbf{K}\,[\,\mathbf{R}\mid\mathbf{t}\,]\,{\mathbf{m}}_i,
    \label{eq:proj_model}
\end{equation}

where $\mathbf{K}$ is the camera intrinsic matrix, $\mathbf{R}\in\mathrm{SO}(3)$ is the rotation matrix, and $\mathbf{t}\in\mathrm{R}(3)$ is the translation, and $\lambda_i$ is the scale factor . The homogenous coordinates of the marker corners are given by ${\mathbf{m}}_i=[X_i,Y_i,0,1]^T$. IPPE solves \eqref{eq:proj_model} and returns two pose solutions $(\mathbf{R}^{(k)},\mathbf{t}^{(k)})$, $k\in\{1,2\}$. The final solution is given by the one with the lowest mean reprojection error obtained as

\begin{equation}
    e^{(k)} = \frac{1}{4}\sum_{i=1}^{4}
      \bigl\|\mathbf{u}_i - \pi\!\bigl(\mathbf{K}\,
      [\mathbf{R}^{(k)}\mid\mathbf{t}^{(k)}]\,{\mathbf{m}}_i\bigr)\bigr\|_2^2,
    \quad k\in\{1,2\},
    \label{eq:reproj}
\end{equation}

where $\pi$ maps homogeneous image coordinates to inhomogeneous (pixel) coordinates. The lower-error hypothesis is taken as the final 6-DoF pose $(\hat{\mathbf{R}},\hat{\mathbf{t}})$. IPPE is faster and more stable than iterative PnP for planar markers because it is non-iterative and free of an initial guess. For non-planar targets on curved surfaces it can provide local approximations.

\begin{figure}
\includegraphics[width=\columnwidth]{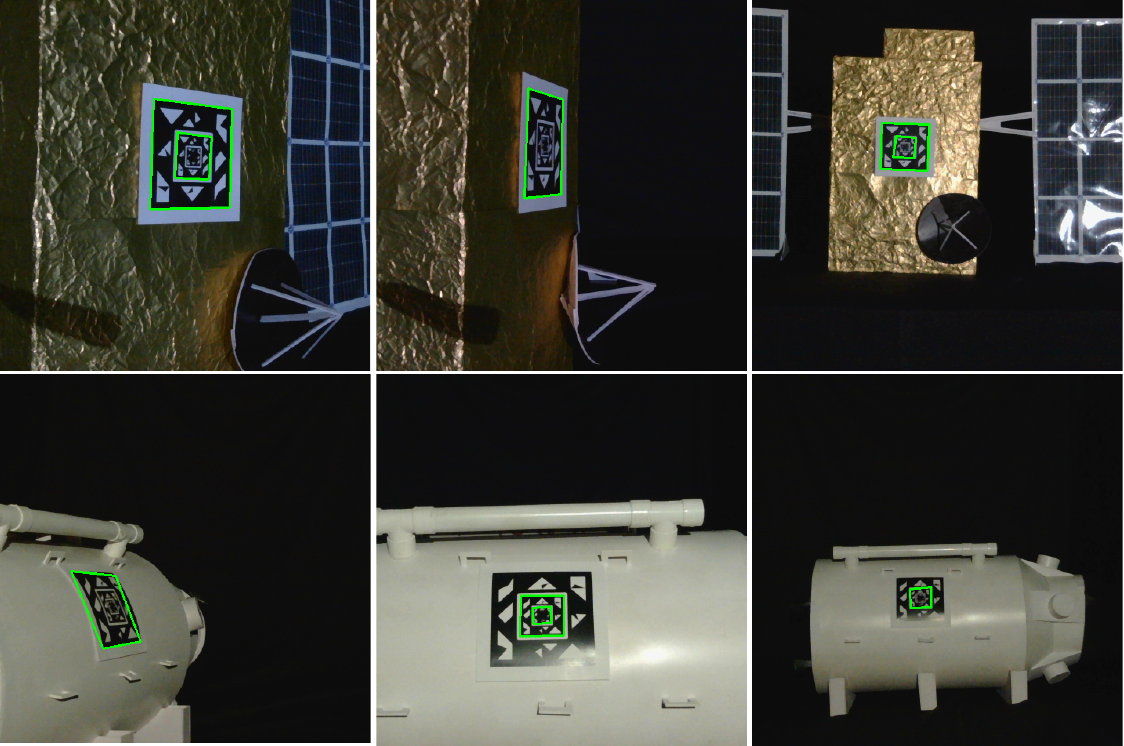}
\caption{Sample AstraTag detections from the validation set. The detected quadrilateral, marker identity, and pose overlay are shown. Top row: spacecraft mockup with marker on golden thermal insulation and solar panels. Bottom row: space station cylindrical module at frontal and significant out-of-plane viewing angles.}
\label{fig:detection}
\end{figure}

\section{EXPERIMENTS AND RESULTS}

\begin{figure*}[t]
    \centering
        \includegraphics[width=\textwidth]{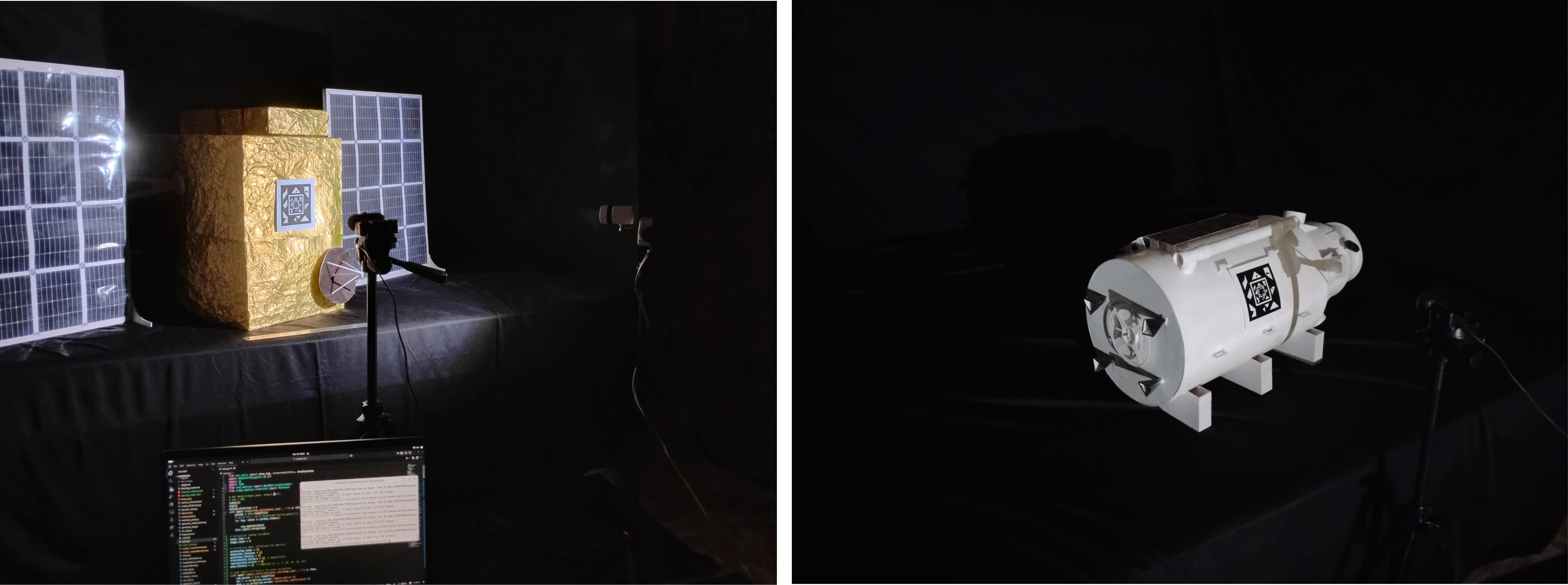}

    \caption{Experimental setup. (a)~Aditya~L1 mockup (flat surface): gold insulation-wrapped bus with solar panels, camera on tripod, point light source on stand, and laptop running the data collection script. (b)~BAS-01 mockup (curved surface): cylindrical module with AstraTag affixed to the curved surface. Both setups use a dark background to approximate space-like illumination.}
    \label{fig:exp_setup}
\end{figure*}

\subsection{Experimental Setup}

The proposed marker was tested using a scaled mockup of spacecrafts under simulated space lighting conditions. The experimental setup consists of a camera mounted on a tripod, connected to a data collection system. The mockups were illuminated using a point light source fixed on a stand at a constant distance for all trials. Both mockups were placed against a dark background to approximate space-like illumination conditions. Figure~\ref{fig:exp_setup} shows both setups. AstraTag was benchmarked against two common visual fiducial systems used in robotics, ArUco\cite{romero2019fractal} and AprilTag\cite{krogius2019flexible}. 

For the experiment, the fractal version of ArUco was used with three recursive layers. A similar version of AprilTag was created by manually by placing. All three markers were specified on a $10 \times 10$\,cm tag size according to their respective conventions and tested under identical camera and lighting conditions.  This nominal specification does not correspond to equal physical print area. AstraTag and ArUco both define the size measured from the vertices of the outermost edge. However, AprilTag defines the size using the four detected corners in the inner region of the marker. The outer layer has additional data cells. This results in AprilTag having more pixel area as compared to other for the same marker dimension. The relative sizes of the markers used in experiment are in Fig~\ref{fig:marker_size}


\subsection{Marker Dataset} 
\begin{figure}[t]
\includegraphics[width=\columnwidth]{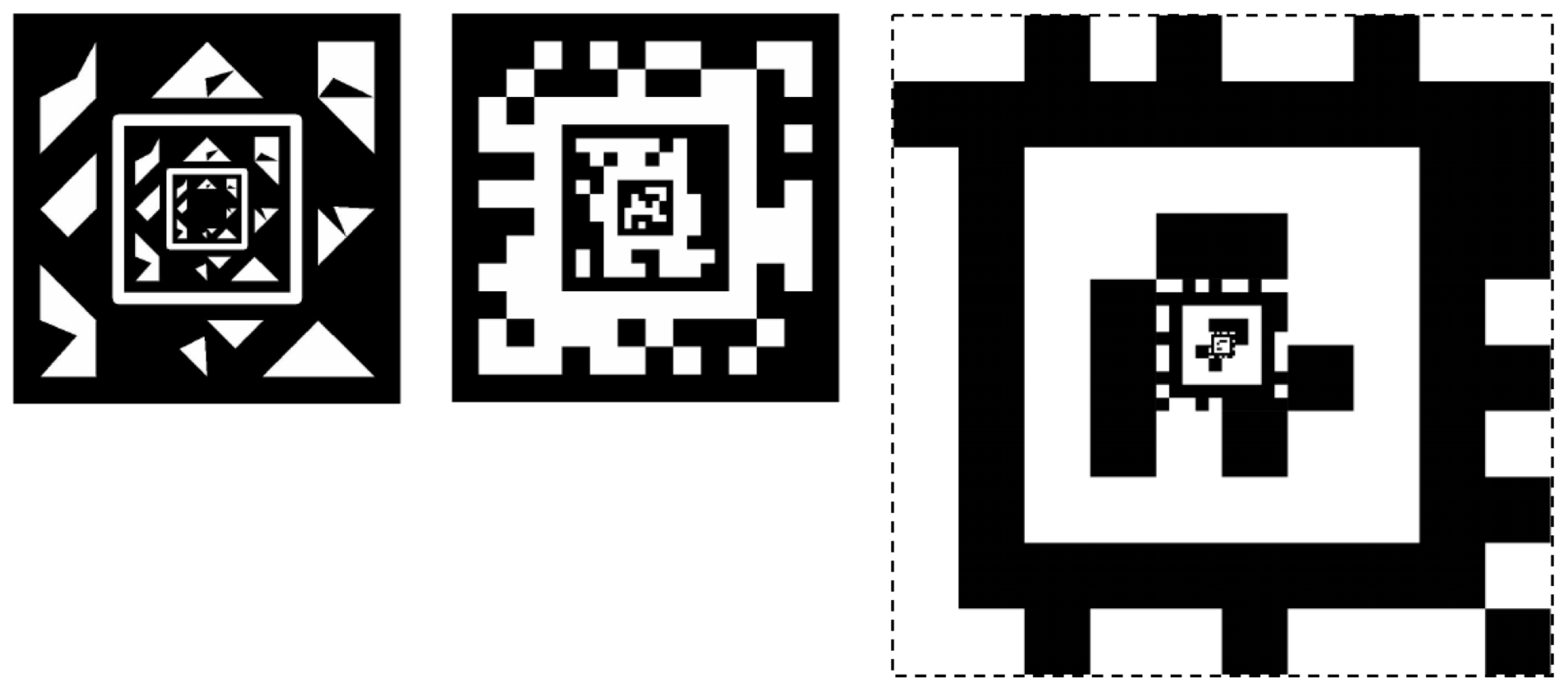}
\caption{The three printed markers used in the benchmark, rendered to true relative scale. Specifying a common $10 \times 10$\,cm tagsize produced AprilTag prints that are $1.67\times$ wider per dimension and $2.76\times$ larger in area than AstraTag and Fractal ArUco. The dashed line in the AprilTag panel indicates the cut boundary of the printed sheet.}
\label{fig:marker_size}
\end{figure}

\begin{figure}
    \centering
    \includegraphics[width=\columnwidth]{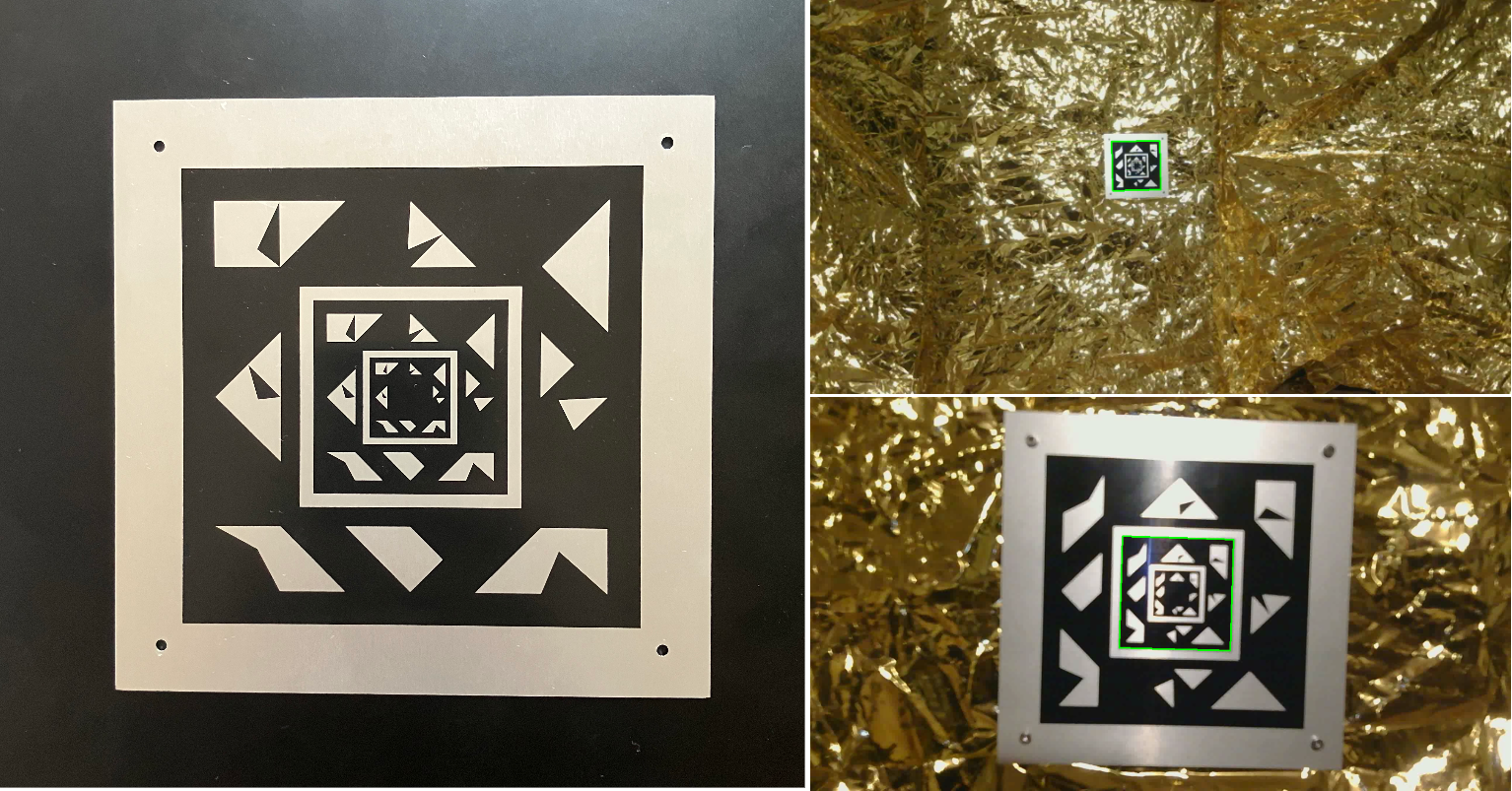}
    \caption{AstraTag marker printed on a Metalphoto anodized aluminum. The two test images shows the marker detection with green bounding box.}
    \label{fig:metalphoto}
\end{figure}

The dataset for evaluation were created to test the performance of the markers on both curved and flat surface. An Space Station dataset was created using cylindrical module mockup inspired by the proposed Bharatiya Antariksh Station module BAS-01. This was used as the curved-surface target. Markers were affixed to the cylindrical surface. The flat-surface target was inspired by the Aditya-L1 coronagraphy mission. The mockup comprises a rectangular bus  wrapped in golden insulation with solar panels. Markers were affixed to the flat face of the bus. The images were collected for target mockups at different distances and angles. An out-of-plane rotation data were also created by placing the camera at fixed distance and rotating the target. We measured the actual printed dimensions of all three cut-out markers used in the experiments. Table~\ref{tab:marker_dims} gives the measurements; Figure~\ref{fig:marker_size} shows the three markers rendered to scale. Moreover, the AstraTag marker was printed with metalphoto anodized aluminum, which can survive extreme temperatures and exposure to Ultraviolet radiation in space environment. The marker pattern is sealed inside anodic oxide layer. Figure~\ref{fig:metalphoto} shows AstraTag printed on Metalphoto.

\begin{table}[ht]
\centering
\begin{tabular}{l|c|c|c}
\toprule
\textbf{Marker} & \textbf{Print dimensions} & \textbf{Print area} & \textbf{Area ratio} \\
                & (mm) & (cm$^2$) & (vs.\ AstraTag) \\
\midrule
AstraTag                       & $100.0 \times 100.0$  & 100.0 & $1.00\times$ \\
Fractal ArUco             & $100.0 \times 100.0$  & 100.0 & $1.00\times$ \\
AprilTag     & $\mathbf{166.6 \times 166.6}$ & $\mathbf{277.5}$ & $\mathbf{2.76\times}$ \\
\bottomrule
\end{tabular}
\caption{Physical printed dimensions of the three markers used in the benchmark, measured directly from the print artwork (cut-line bounding box). All three were specified at a $10 \times 10$\,cm tagsize, but the AprilTag tagCustom48h12 layout extends data cells outside the tagsize boundary, producing a print that is $1.67\times$ wider in each linear dimension and $2.76\times$ larger in surface area than AstraTag or Fractal ArUco.}
\label{tab:marker_dims}
\end{table}

\subsection{Results}
\label{sec:results}

\subsubsection{Space Station Dataset}

AstraTag maintains detection rate between 97--100\% when the target rotated from 10\textdegree{} through 70\textdegree{} on the curved surface. This is enabled by the TPS fallback option that compensates for the combined perspective and cylindrical distortion. Fractal ArUco achieves 93--98\% at moderate angles but collapses to 2\% at 40\textdegree{} and 0\% when the rotation increases further, as its square-grid inner layers cannot be resolved under the compounded curvature distortion. AprilTag sustains 100\% detection up to 60\textdegree{} but drops to 56\% at 70\textdegree{}. At 80\textdegree{}, all three markers fail, the marker is nearly on the edge. This results in contour-based detector failing to recover quadrilateral from the marker boundary.

\begin{figure}[t]
    \centering
    \begin{tikzpicture}
    \begin{groupplot}[draxis,
        group style={group size=1 by 2, vertical sep=2.3cm},
        width=0.8\linewidth, height=5.0cm]
    \nextgroupplot[title={(a) Distance}, xlabel={Distance (cm)},
        ylabel={Detection Rate (\%)}, xtick={30,50,70,90,110,130,150}]
        \addplot[aruco]    coordinates {(30,24)(50,84)(70,29)(90,16)(110,0)(130,0)(150,0)};
        \addplot[astratag] coordinates {(30,100)(50,100)(70,78)(90,100)(110,93)(130,90)(150,49.49)};
        \addplot[apriltag] coordinates {(30,100)(50,100)(70,100)(90,100)(110,100)(130,100)(150,100)};
    \nextgroupplot[title={(b) Out-of-plane rotation}, xlabel={Rotation (deg)},
        ylabel={Detection Rate (\%)}, xtick={10,20,30,40,50,60,70,80},
        legend style={font=\small, fill=white, fill opacity=0.9, text opacity=1,
            draw=gray!60, row sep=-1pt, inner sep=2pt,
            at={(0.03,0.04)}, anchor=south west}]
        \addplot[aruco]    coordinates {(10,98)(20,93)(30,95)(40,2)(50,0)(60,0)(70,0)(80,0)};
        \addplot[astratag] coordinates {(10,100)(20,100)(30,100)(40,100)(50,100)(60,100)(70,97)(80,0)};
        \addplot[apriltag] coordinates {(10,100)(20,100)(30,100)(40,100)(50,100)(60,100)(70,56)(80,0)};
        \legend{Fractal ArUco (3L), AstraTag, AprilTag (48h12)}
    \end{groupplot}
    \end{tikzpicture}
    \caption{Detection rate comparison on the space station dataset (curved surface). (a)~Detection rate versus distance. (b)~Detection rate versus out-of-plane rotation angle. AstraTag and AprilTag maintain high detection rates across distances, while AstraTag's TPS fallback provides rotation robustness on the curved surface.}
    \label{fig:benchmark}
\end{figure}
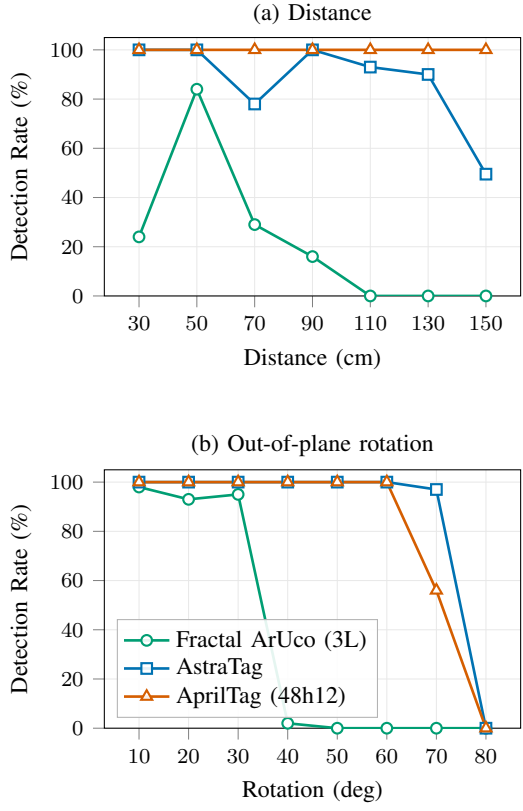

When tested in multiple ranges, AprilTag achieves 100\% detection at all distances. This could be attributed to its relatively large nominal size and optimized quad detector. AstraTag averages 87.2\%, with a drop at 150\,cm where the recursive inner layers become too small to resolve on the curved surface. Fractal ArUco averages only 21.9\%, with detection limited to 50\,cm and below. The inner fractal layers of the ArUco require substantially more pixels to decode than the outer layer, and the cylindrical distortion further degrades their visibility. Figure~\ref{fig:benchmark} summarizes the detection rate as a function of distance and rotation angle for all three markers on the Space Station dataset.

\subsubsection{Spacecraft Dataset}

To isolate the effect of surface curvature, the same three markers were tested on a flat surface of the Aditya-L1 spacecraft mockup. On the flat surface, all three markers perform substantially better. Compared to 21.9\% on the curved surface, Fractal ArUco recovers to a mean detection rate of 99.3\% on the flat surface over multiple distances. It shows 89.5\% detection on rotation compared to 36.0\% on the curved surface, with its only failure at 75\textdegree{} rotation when detection rate falls 58\%. AstraTag and AprilTag achieve 100\% detection in all conditions on the flat surface target. This demonstrates that the performance gap observed on the space station dataset is primarily due to failure in detecting outer quadrilateral. All three fiducial markers have a lower detection rate in the curved surface dataset. 

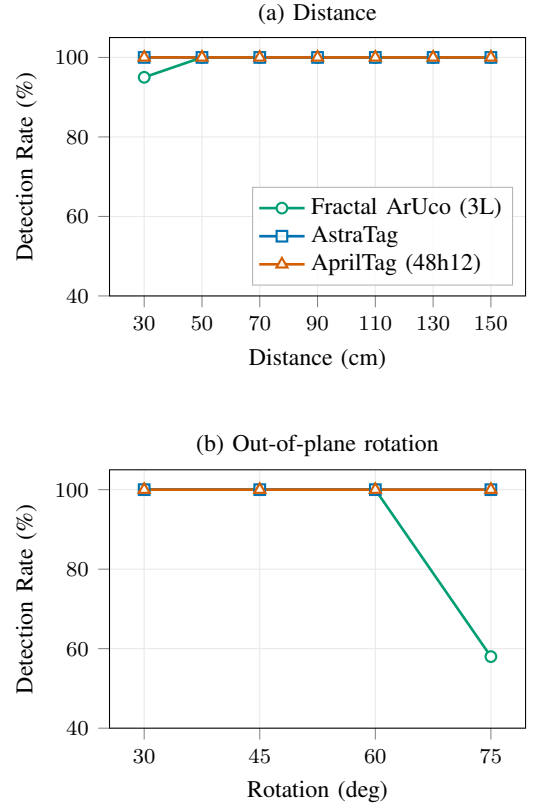
\begin{figure}[t]
    \centering
    \begin{tikzpicture}
    \begin{groupplot}[draxis,
        group style={group size=1 by 2, vertical sep=2.3cm},
        width=0.8\linewidth, height=5.0cm, ymin=40, ytick={40,60,80,100}]
    \nextgroupplot[title={(a) Distance}, xlabel={Distance (cm)},
        ylabel={Detection Rate (\%)}, xtick={30,50,70,90,110,130,150},
        legend style={font=\small, fill=white, fill opacity=0.9, text opacity=1,
            draw=gray!60, row sep=-1pt, inner sep=2pt,
            at={(0.97,0.04)}, anchor=south east}]
        \addplot[aruco]    coordinates {(30,95)(50,100)(70,100)(90,100)(110,100)(130,100)(150,100)};
        \addplot[astratag] coordinates {(30,100)(50,100)(70,100)(90,100)(110,100)(130,100)(150,100)};
        \addplot[apriltag] coordinates {(30,100)(50,100)(70,100)(90,100)(110,100)(130,100)(150,100)};
        \legend{Fractal ArUco (3L), AstraTag, AprilTag (48h12)}
    \nextgroupplot[title={(b) Out-of-plane rotation}, xlabel={Rotation (deg)},
        ylabel={Detection Rate (\%)}, xtick={30,45,60,75}]
        \addplot[aruco]    coordinates {(30,100)(45,100)(60,100)(75,58)};
        \addplot[astratag] coordinates {(30,100)(45,100)(60,100)(75,100)};
        \addplot[apriltag] coordinates {(30,100)(45,100)(60,100)(75,100)};
    \end{groupplot}
    \end{tikzpicture}
    \caption{Detection rate comparison on the Aditya~L1 dataset (flat surface). All three markers achieve near-perfect detection, confirming that surface curvature is the primary differentiator.}
    \label{fig:benchmark_aditya}
\end{figure}

\begin{figure}[t]
    \centering
    \begin{tikzpicture}
    \begin{axis}[draxis, width=0.8\linewidth, height=5.0cm,
        xlabel={Distance (cm)}, ylabel={Detection Rate (\%)},
        xtick={30,50,70,90,110,130,150},
        legend style={at={(0.03,0.03)}, anchor=south west, font=\footnotesize,
            fill=white, draw=gray!60}]
        \addplot[tauA] coordinates {(30,100)(50,100)(70,100)(90,100)(110,100)(130,100)(150,100)};
        \addplot[tauB] coordinates {(30,100)(50,100)(70,100)(90,100)(110,100)(130,100)(150,100)};
        \addplot[tauC] coordinates {(30,0)(50,100)(70,100)(90,100)(110,100)(130,100)(150,100)};
        \legend{$\tau_H=10$, $\tau_H=5$, $\tau_H=3$}
    \end{axis}
    \end{tikzpicture}
    \caption{Detection rate of the Metalphoto AstraTag taken at $1280\times720$ resolution, at various Hamming thresholds}
    \label{fig:metalphoto_detection}
\end{figure}
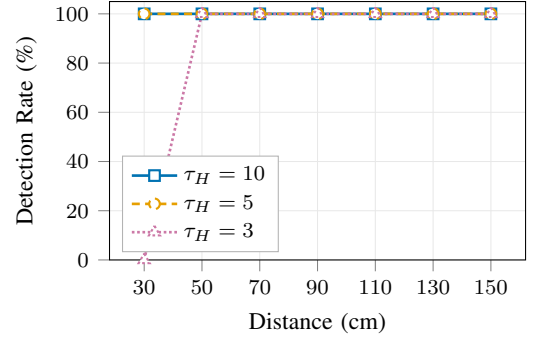

\subsection{Discussion}
This relative difference in nominal size of the AprilTag means that at any given camera distance, it occupies approximately two and half times as many image pixels as the other two markers, which has different implications for different parts of the evaluation process. The evaluation of markers across range of distances on both curved and flat surfaces, AprilTag has more pixel coverage. Thus, at the standard tagsize specification, AprilTag's larger physical print produces more pixels per cell in the image making it detectable at distances where other markers fail. To assess the algorithmic efficiency for detection, an experiment with equal area for all the three markers is required. AstraTag demonstrates detection at higher incidence angle due to \ac{TPS} fallback option. Figure~\ref{fig:detection} shows sample detections from the test set across both the spacecraft mockup and the space station module. For the Metalphoto version of AstraTag, the detection and identification are $100\%$ for $\tau_H\le5$ across the entire range; the strict $\tau_H=3$ fails only at $30$\,cm, where the satin-finish specular glare reduces the signature margin, where $\tau_H$ is the Hamming distance(see Figure~\ref{fig:metalphoto_detection}). Figure~\ref{fig:multi_range} shows the marker detection results for images taken with low resolution camera.
    
\section{CONCLUSIONS AND FUTURE WORK}

This paper presented AstraTag, a recursive fiducial marker based on the Spidron fractal pattern, designed for autonomous on-orbit robotic operations. The marker uses a fixed triangular subdivision of Spidron regions for encoding, with each region subdivided into six triangles via its vertices, edge midpoints, and centroid. The resulting 48-bit signature is constructed from a \ac{GRS} code which allows to design a marker with configurable payload. The current implementation performs identification via dictionary matching. This approach will be replaced by algebraic \ac{GRS} decoding in future work. The detection pipeline combines contour-based quadrilateral localisation with \ac{TPS} warping as a curved surface fallback, using the marker's internal rectangular borders as geometric correspondences. Comparative benchmarks against Fractal ArUco and AprilTag  under simulated conditions on a curved space station mockup demonstrate that AstraTag achieves better detection rate across out-of-plane rotation angles on cylindrical surfaces. On a flat-surface dataset, all three markers achieve near-perfect detection, confirming that surface curvature is the primary differentiator. The self similar pattern of AstraTag makes it suitable for proximity operations and docking for space robots. The marker reduces the computational cost involved in on-orbit operations not limited to spacecraft servicing, in-space assembly, and manufacturing.



\bibliographystyle{IEEEtran}   
\bibliography{ref}

\end{document}